\documentclass[runningheads]{llncs}

\usepackage{kotex}
\usepackage{enumitem}
\usepackage{accv}

\usepackage{accvabbrv}

\usepackage{graphicx}
\usepackage{booktabs}

\usepackage[accsupp]{axessibility}  

\newcommand\highlight[1][yellow]{%
  \bgroup
  \markoverwith{\textcolor{#1}{\vrule width.1em height.8em depth.2em}}%
  \ULon
}

\usepackage{hyperref}

\usepackage{orcidlink}

\usepackage{multirow}
\usepackage{adjustbox}
\begin{document}

\title{Octave-YOLO: Cross frequency detection network with octave convolution} 

\titlerunning{Abbreviated paper title}

\author{Sangjune Shin,
Dongkun Shin$\dagger$ }

\authorrunning{F.~Author et al.}

\institute{Department of Electrical and Computer Engineering, Sungkyunkwan University 
\email{\{sangjune97,dongkun\}@skku.edu}\\
}

\maketitle

\begin{abstract}

Despite the rapid advancement of object detection algorithms, processing high-resolution images on embedded devices remains a significant challenge. Theoretically, the fully convolutional network architecture used in current real-time object detectors can handle all input resolutions. However, the substantial computational demands required to process high-resolution images render them impractical for real-time applications. To address this issue, real-time object detection models typically downsample the input image for inference, leading to a loss of detail and decreased accuracy.

In response, we developed Octave-YOLO, designed to process high-resolution images in real-time within the constraints of embedded systems. We achieved this through the introduction of the cross frequency partial network (CFPNet), which divides the input feature map into low-resolution, low-frequency, and high-resolution, high-frequency sections. This configuration enables complex operations such as convolution bottlenecks and self-attention to be conducted exclusively on low-resolution feature maps while simultaneously preserving the details in high-resolution maps. Notably, this approach not only dramatically reduces the computational demands of convolution tasks but also allows for the integration of attention modules, which are typically challenging to implement in real-time applications, with minimal additional cost. Additionally, we have incorporated depthwise separable convolution into the core building blocks and downsampling layers to further decrease latency.

Experimental results have shown that Octave-YOLO matches the performance of YOLOv8 while significantly reducing computational demands. For example, in 1080x1080 resolution, Octave-YOLO-N is 1.56 times faster than YOLOv8, achieving nearly the same accuracy on the COCO dataset with approximately 40 percent fewer parameters and FLOPs.
\end{abstract}

\section{Introduction}
\label{sec:intro}
\begin{figure}[tp]
  \centering
  \begin{subfigure}{0.475\linewidth}
  \centering
  \includegraphics[width=6cm]{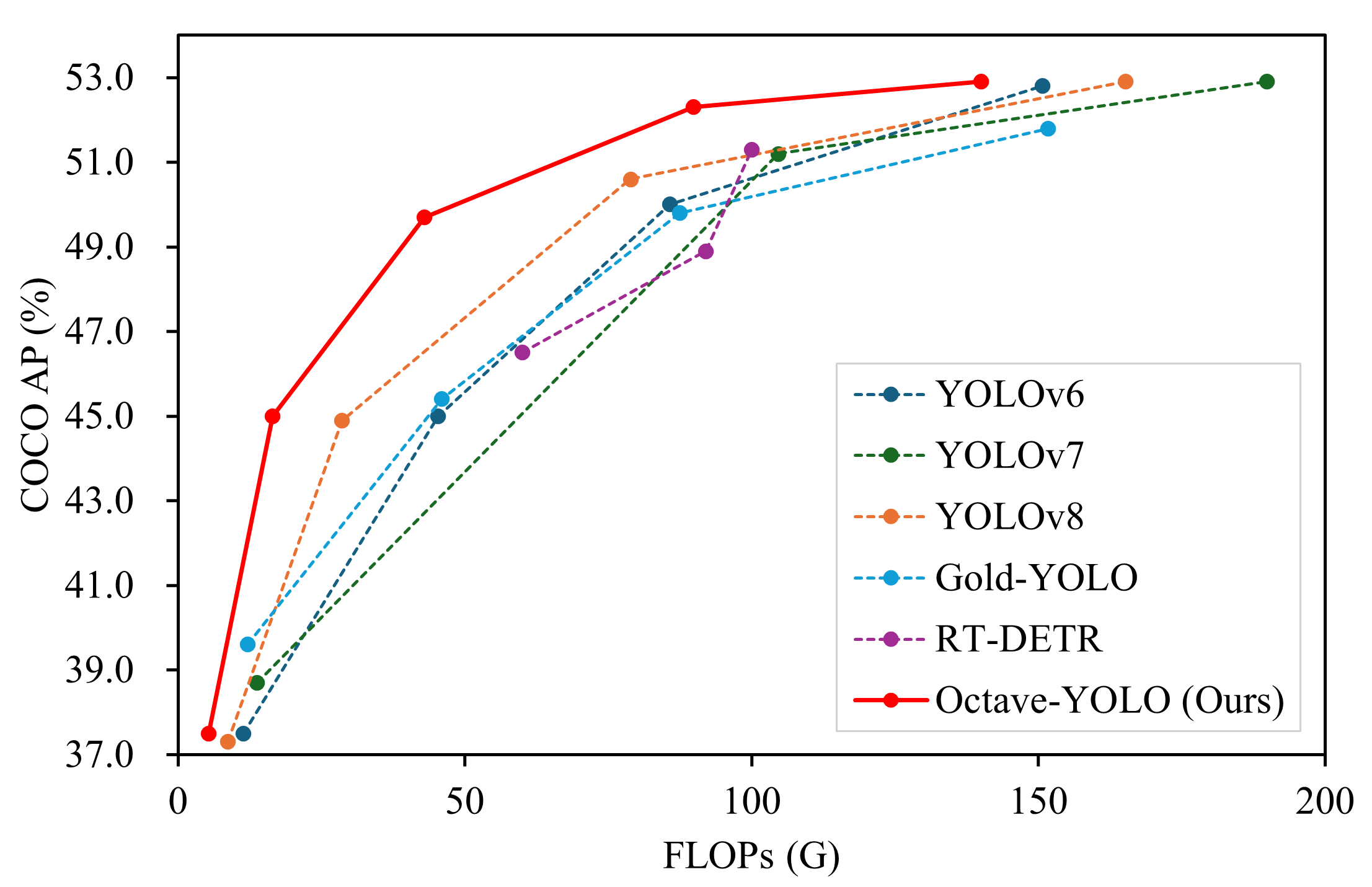}
  \end{subfigure}
  \hfill
  \begin{subfigure}{0.475\linewidth}
  \centering
  \includegraphics[width=6cm]{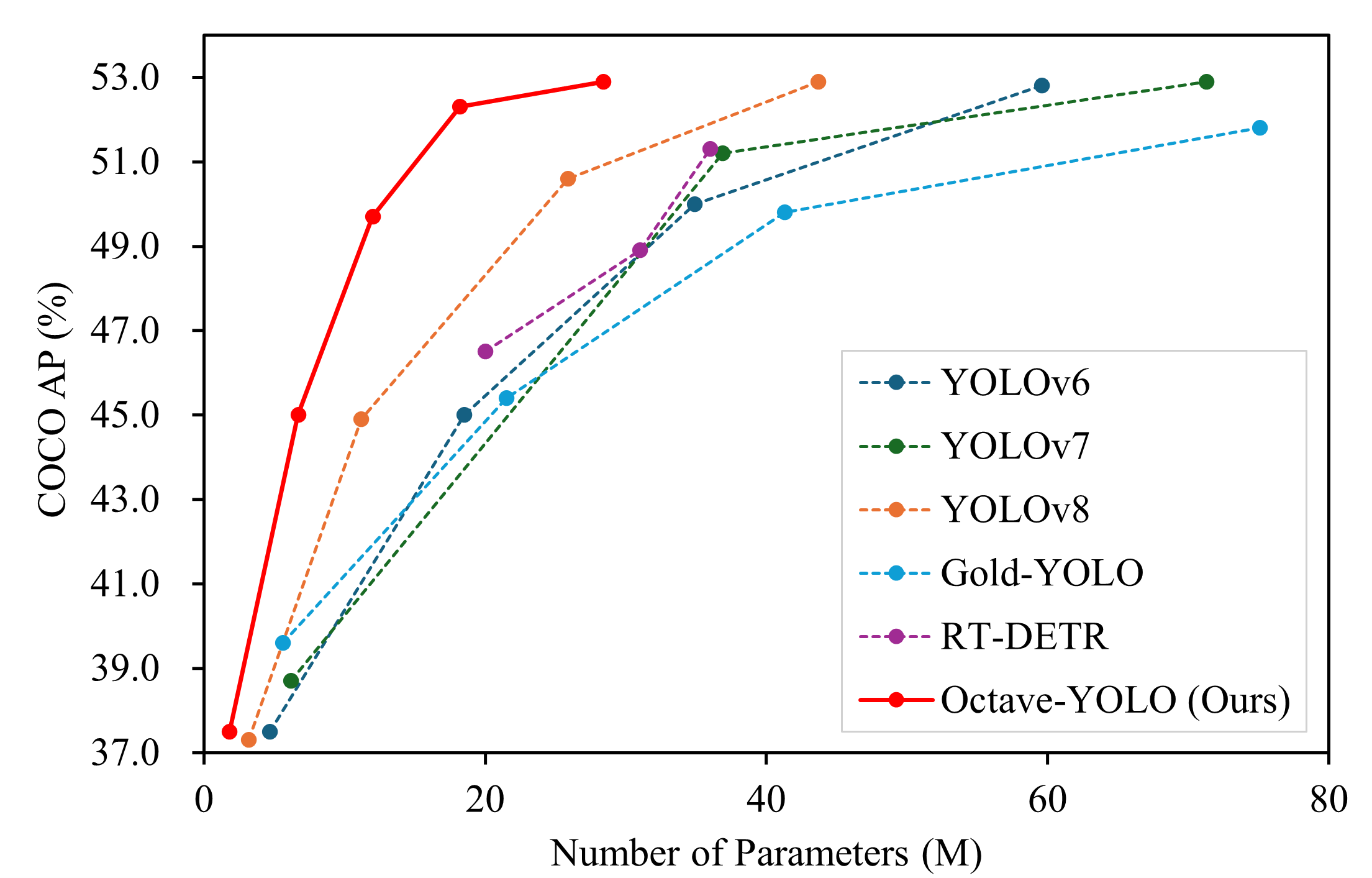}
  \end{subfigure}
  \caption{Comparisons with others in terms of FLOPs vs AP (left) and model size vs AP (right) trade-offs.}
  \label{fig:fig1}
\end{figure}
Real-time object detection plays a crucial role in the field of computer vision, aiming to accurately and swiftly identify the types and locations of objects within images. This technology is extensively utilized in various practical applications such as autonomous driving, robotics, object tracking, drones, and UAVs. Recently, the advent of deep learning algorithms has significantly improved the ability of object detection systems to learn complex patterns within images, thereby enhancing the accuracy and robustness of detections. Despite the ease of acquiring high-resolution images on embedded devices that utilize object detection, the use of deep learning for real-time detection still necessitates the downsampling of original high-resolution images to lower resolutions such as 640x640 or 512x512 due to computational and memory constraints. This results in the loss of detailed information and reduces the capability to detect smaller objects, adversely affecting overall accuracy.

This research aims to overcome these limitations and enhance the inference speed of YOLO in high-resolution settings within constrained embedded environments. We propose the cross frequency partial network (CFPNet) structure, which processes the input feature map by splitting it into high-frequency, high-resolution parts and low-frequency, low-resolution parts. This approach enables the model to perform complex computations in the low-frequency, low-resolution domain during inference and training while simultaneously preserving high-resolution information, achieving competitive performance with high efficiency. Additionally, we actively incorporate depthwise separable convolution to reduce computational redundancy and further decrease latency.\\\\Our specific contributions include:\\
\begin{itemize}[leftmargin=*,topsep=-3pt,itemsep=0pt]
\item\textit{Frequency Separable Block (FSB).} An enhancement of YOLOv8's C2f module, the FSB utilizes the CFPNet to segregate the input feature map into high and low frequency components. This block processes only the low-frequency portion through the convolution bottleneck, optimizing convolution efficiency by integrating depthwise separable convolution for both spatial and channel separation.
\item\textit{Frequency Separable Self-Attention (FSSA).} To improve efficiency, we replace traditional self-attention mechanisms, typically heavy on computation and memory, with our FSSA module. By splitting the feature map through CFPNet, only the low-resolution, low-frequency components are subjected to self-attention processing. This modification effectively integrates global representation capabilities into YOLO at minimal additional costs, thus enhancing model performance.
\item\textit{Depthwise Separable Downsampling.} Replacing standard 3x3 convolutions with depthwise separable convolutions, we delineate tasks of resolution reduction and channel increase, preserving information during downsampling and reducing latency.\\
\end{itemize}
Employing these methodologies, we have successfully developed a new lineup of real-time end-to-end object detectors, designated as Octave-YOLO-N / S / M / B / L / X. As shown in \cref{fig:fig1}, extensive testing on benchmarks like the COCO dataset reveals that Octave-YOLO substantially surpasses previous models in computation-accuracy trade-offs across different scales. Notably, our Octave-YOLO-N/S models maintain slightly better performance compared to YOLOv8-N/S, while being about 1.56 times faster in high-resolution latency. Additionally, the larger Octave-YOLO-X model has 58.35\% fewer parameters and 45.65\% fewer operations compared to YOLOv8-X.

\section{Related Work}
\subsection{Frequency Domain Vison Tasks}

In the field of computer vision, approaches like multi-scale feature fusion within the frequency domain have been extensively researched. Before the advent of deep learning, multi-scale representations were already employed for local feature extraction, as seen with SIFT features. In the deep learning era, these representations continue to play a critical role due to their robustness and capacity for generalization.

Frameworks like FPN\cite{Lin_2017_CVPR} and PSP\cite{richardson2021encoding} integrate convolutional features from various depths at the network's end for object detection and segmentation tasks. MSDNet\cite{huang2017multi} and HR-Nets\cite{sun2019deep} introduced network architectures with multiple branches, each branch maintaining its own spatial resolution, thus allowing for diverse scale processing.

Xu et al.\cite{xu2020learning} proposed a novel approach to learning computer vision tasks in the frequency domain. They utilized the discrete cosine transform (DCT) of the input image as the input to a convolutional network. This method demonstrated satisfactory results in computer vision tasks, maintaining a slight decrease in accuracy even when most parameters were pruned. Zhong et al.\cite{zhong2022detecting} achieved high accuracy in object detection by integrating both frequency and pixel domain information, enhancing the detection capabilities. Cai et al.\cite{cai2018cascade} developed a robust method for image translation by decomposing GAN\cite{goodfellow2020generative}-based image-to-image translation into low and high-frequency components and processing them within the same frequency domain. 

Additionally, the Multiscale Vision Transformers\cite{fan2021multiscale} demonstrated their effectiveness in the Vision Transformer (ViT) architecture by adapting to images of varying scales. This was achieved through the multi-head pooling attention module, which can handle inputs of different scales via a pooling layer, thereby enhancing the model’s adaptability and performance across different image resolutions.

\begin{figure}
    \centering
    \includegraphics[height=4cm]{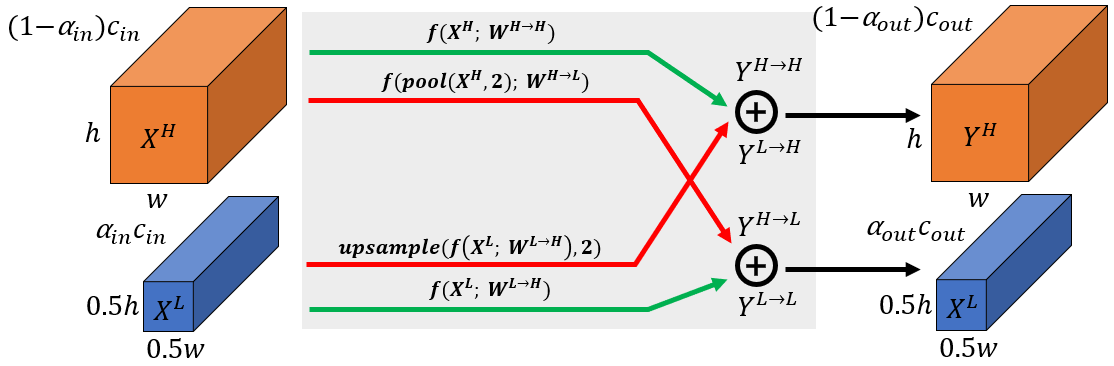}
    \caption{The detailed structure of the octave convolution. The two green paths represent the updating of information for the high and low frequency feature maps, respectively, while the two red paths represent the mutual exchange of information between the two different frequencies.}
    \label{fig:OctConv}
\end{figure}

Octave convolution\cite{chen2019drop}, emerging to enhance the efficiency and accuracy of computer vision models, can extract features of various frequencies while reducing computational costs. Unlike standard convolution, octave convolution accounts for high-frequency and low-frequency components in the input and output feature maps of a convolutional network, as shown in \cref{fig:OctConv}. Low-frequency components capture the overall shape and structural features of an object but are often redundant. High-frequency components are used to capture the edges and detailed textures. In octave convolution, low-frequency components refer to the feature map obtained through pooling, while high-frequency components refer to the original feature map without pooling. Due to the redundancy of low-frequency components, their feature map size is set to half that of the high-frequency components. Octave convolution can replace traditional convolution operations, typically achieving higher performance while reducing the amount of computation because the low-frequency feature map has half the resolution of the original. Notably, despite the reduced resolution of the low-frequency feature maps, using convolutional kernels of the same size as vanilla convolution effectively doubles the receptive field.

In addition to significant savings in computation and memory, octave convolution enhances recognition performance by enlarging the receptive field size, which facilitates effective communication between high and low frequencies and captures more global information.

In octave convolution, high-frequency and low-frequency information is extracted from the input, and these components are fused to create new high-frequency and low-frequency outputs. Hyperparameters such as $\alpha_{in}$ and $\alpha_{out}$ are adjusted to control the ratio of high-frequency and low-frequency feature maps in both the input and output. Setting these parameters to 0 or 1 allows the network to receive only high-frequency inputs or produce only low-frequency outputs, respectively. This flexibility in parameter setting enables fine-tuning of the convolution process to optimize for specific tasks or computational efficiency.

\subsection{Object Detection}

General object detection aims to localize a predefined object in a given input image and simultaneously classify the object's class. Nowadays, various deep learning-based approaches have been proposed to solve the object detection problem. These approaches can be broadly categorized into two categories: two-stage detectors and one-stage detectors. 

The two-stage detector is mainly based on R-CNN\cite{girshick2014rich}, which extracts regions through region proposal in the first stage, and the extracted regions are assigned to various classes based on their inferred probability values. Later, more efficient models have been proposed based on R-CNN\cite{girshick2014rich}, such as fast R-CNN\cite{girshick2015fast}, faster R-CNN\cite{ren2015faster}, mask R-CNN\cite{he2017mask}, and cascade R-CNN\cite{cai2018cascade}. However, while these two-stage detectors perform very well, they have too long latency and slow speed. 

To overcome these problems, one-stage detectors have been studied. Common one-stage detectors include SSD\cite{liu2016ssd}, YOLO series\cite{Jocher_Ultralytics_YOLO_2023,li2022yolov6, Jocher_YOLOv5_by_Ultralytics_2020, bochkovskiy2020yolov4, wang2023yolov7 }, Retina-Net\cite{lin2017focal}, \etc. They do not separate region proposal and classification, but treat them as a single regression problem.

In addition to CNN-based architectures, recent research has also seen the emergence of Transformer-based architectures. Transformer-based object detectors (DETRs)\cite{carion2020end} have simplified the object detection pipeline by removing hand-crafted components such as non-maximum suppression (NMS) and realized end-to-end object detection. However, the high computational cost of DETRs makes it difficult to implement real-time object detection.
Although some real-time object detectors have been proposed, such as RT-DETR\cite{lv2023detrs}, the DETR series object detectors are very difficult to apply to new domains without a pre-trained model of that domain, so the most widely used real-time object detectors are still the YOLO series.
\subsection{YOLO Series}

The YOLO series, a representative one-stage object detector, was first published in 2015\cite{redmon2016you}. It uses a single neural network to perform all the necessary steps in the object detection problem. As a result, it achieves not only very good detection performance, but also real-time speed. YOLOv4\cite{bochkovskiy2020yolov4} introduced CSP network\cite{wang2020cspnet} in the backbone structure to improve the learning ability of the CNN, allowing the network to be lightweight while maintaining the accuracy of feature map estimation. In addition, PAFPN\cite{8579011} module was introduced to the network's neck structure.

YOLOv5\cite{Jocher_YOLOv5_by_Ultralytics_2020} introduced the focus module in the starting structure of the backbone to speed up the processing of input images. YOLOX\cite{ge2021yolox} still uses CSPDarkNet as its backbone network, but introduced SiLU\cite{elfwing2018sigmoid} instead of ReLU\cite{glorot2011deep} to solve the problem of gradient variance. The authors also introduced a decoupled head that implements confidence and box regression prediction separately, and implemented an anchor-free detector head with SimOTA. YOLOv6\cite{li2022yolov6} introduced EfficientRep\cite{weng2023efficientrep} as the backbone of the network. YOLOv7\cite{wang2023yolov7} introduced the E-ELAN structure to the backbone network for efficient multi-scale feature training. The authors incorporated the E-ELAN structure into the backbone of their study, facilitating effective learning and convergence of the network through the management of both the shortest and longest gradient paths. Jocher \textit{et~al.}~\cite{Jocher_Ultralytics_YOLO_2023} proposed YOLOv8. The authors introduced the C2f module as a backbone configuration module. YOLOv9\cite{wang2024yolov9} proposes GELAN to improve the architecture and introduces PGI to augment the training process. YOLOv10\cite{wang2024yolov10} suggests consistent dual assignments for NMS-free training to achieve efficient end-to-end detection. These YOLO series utilize CSPNet or ELAN and their variations as the main computing units.

\subsection{Cross stage partial network (CSPNet)}
Wang et al. introduced a new CNN backbone structure, the cross stage partial network (CSPNet)\cite{wang2020cspnet}, designed for efficient vision tasks. The main goal of CSPNet is to enhance the model's learning capability while reducing unnecessary redundant computations. As illustrated in \cref{fig:CFPNet}.(a), this is achieved by splitting the feature map of the base layer into two parts and processing each part differently to divide the gradient flow. CSPNet enables a richer combination of gradients by partially integrating and transitioning feature maps between layers. This architecture is particularly well-suited for real-world applications that require fast inference times on devices with limited computational resources. It evenly distributes computational loads across the network's layers and reduces memory costs. In fact, the feature pyramids generated through CSPNet significantly reduce memory usage, saving on the manufacturing costs and space required for expensive DRAM. Additionally, CSPNet prevents the duplication of gradient information during training, ensuring the diversity of gradient flow, which in turn allows the network to learn more efficiently. This ultimately improves both the accuracy and the inference speed of the model. In this paper, we use octave convolution to separate and integrate the feature map into high and low resolution frequency domains.

\section{Methodology}
\subsection{Cross frequency partial network (CFPNet)}
\begin{figure}[t]
    \centering
    \includegraphics[height=8cm]{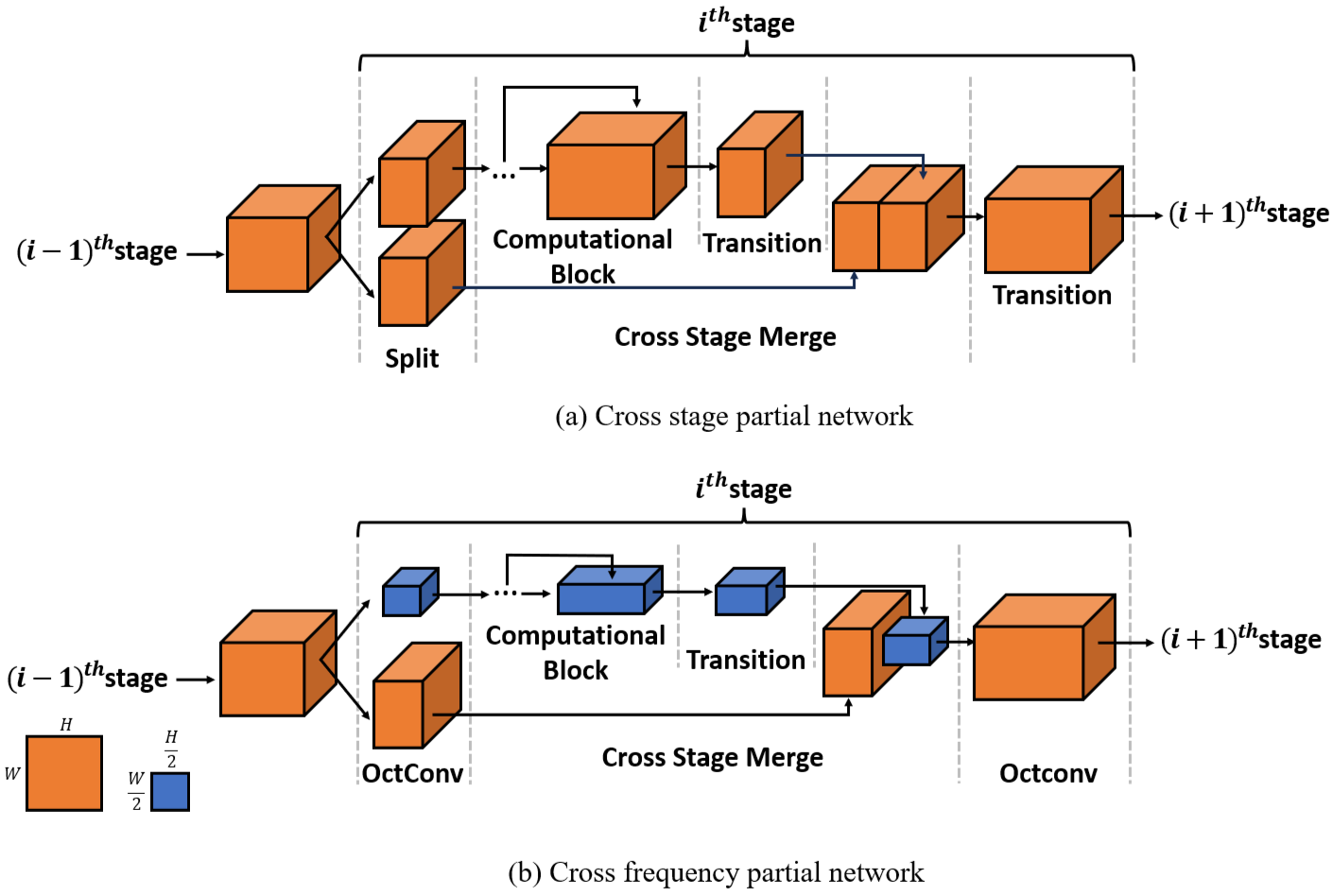}
    \caption{Comparison between the original cross stage partial network (CSPNet) and our proposed cross frequency partial network (CFPNet).}
    \label{fig:CFPNet}
\end{figure}
We first draw inspiration from octave convolution and CSPNet\cite{wang2020cspnet} to propose a cross frequency partial network(CFPNet) structure, which splits the feature map into high-frequency and low-frequency components and processes them differently to reduce redundant calculations and enhance efficiency. The conventional CSPNet, as shown in \cref{fig:CFPNet}.(a), divides the feature map of the input layer into two parts along the channel axis. One part goes through a computation block and the other is combined with the processed feature map to move to the next stage. This computation block could be any computational blocks such as convolution blocks, attention blocks, or residual blocks. The other part traverses the entire stage directly and then integrates with the part that has passed through the computation block. Since only a part of the feature map enters the computation block for processing, this design effectively reduces the amount of design variables, operations, memory traffic, and peak memory, enabling the system to achieve faster inference speeds.
We enhance this efficient CSPNet by integrating Octave convolution, splitting and combining the feature map along the frequency axis instead of the channel axis. As shown in \cref{fig:CFPNet}.(b), we replace the entire process of splitting and combining feature maps in the conventional CSPNet with Octave convolution. Octave convolution allows easy control over the ratio of high-frequency and low-frequency feature maps in the input and output through the parameters $\alpha_{in}$ and $\alpha_{out}$.

Specifically, we first split the feature map of the input layer into high-frequency and low-frequency feature maps using an octave convolution with alphain set to zero. Subsequently, the low-frequency part passes through the computation block and combines with the processed low-frequency feature map to proceed to the next stage. The high-frequency part traverses the entire stage directly without undergoing complex computation processes and integrates with the low-frequency part that has passed through the computation block. During this process, the low-frequency and high-frequency feature maps are fused across frequencies using an octave convolution with alphaout set to zero, restoring the original resolution of the input layer with a vanilla feature representation that does not have multiple frequencies.

Our approach not only involves passing only a part of the feature map but also achieves better efficiency than the conventional CSPNet because the low-frequency feature map with half the resolution of the high-frequency part passes through the computation block. Additionally, by enlarging the receptive field size, it contributes to effective communication between low and high frequencies and captures more global information, which can enhance recognition performance.
We apply this CFPNet structure to the building blocks and self-attention modules of YOLO, achieving higher efficiency without compromising performance.

\begin{figure}[tp]
  \centering
  \includegraphics[width=0.9\textwidth]{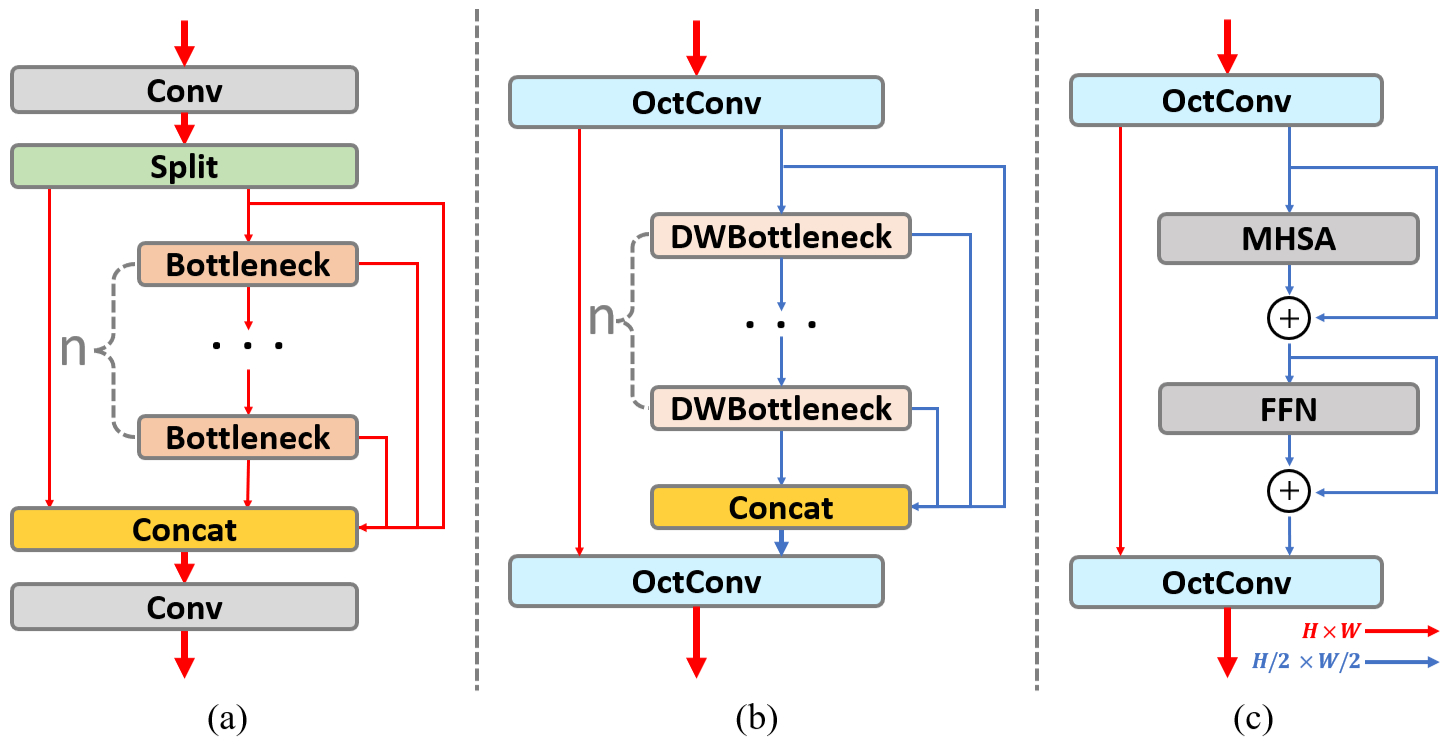}
  \caption{(a) The original C2f building block used in YOLOv8. (b) The frequency serparable block (FSB). (c) The frequency separable self-attention module (FSSA).}
  \label{fig:architecture}
\end{figure}

\subsection{Frequency separable block (FSB)}
YOLO typically extracts features and learns patterns from the input feature maps through convolutional building blocks of the same structure in all stages except the stem and downsampling layers. The C3 block introduced in YOLOv5, which incorpataes CSPNet and the C2f block(\cref{fig:architecture}.(a)) presented in YOLOv8, which employs ELAN\cite{zhang2022efficient} were both efforts towards effective feature extraction and fusion. They aim to improve performance and computational efficiency simultaneously by splitting the input feature map into two parts and processing each part differently through diversified gradient paths. However, these designs only considered separation at the channel level and did not address multi-scale feature representation. We apply the previously introduced CFPNet to design an efficient base building block. We propose a frequency separable block (FSB), as shown in \cref{fig:architecture}.(b), an enhancement of the C2f module from YOLOv8. Similar to the C2f module from YOLOv8, but instead of a split, we use octave convolution to divide the input feature map into high-frequency and low-frequency parts, allowing only the low-frequency part to pass through the convolution bottleneck. Additionally, inspired by Xception\cite{chollet2017xception}, we introduce depthwise separable convolution into the bottleneck for more efficient and faster computations, separating spatial mixing with depthwise convolution and channel mixing with pointwise convolution. Finally, we use octave convolution again to fuse the processed low-frequency part with the preserved high-frequency part. As a result, we implement the FSB, an efficient building block used across the model, achieving very high efficiency with minimal performance degradation.

\subsection{Frequency separable self-attention (FSSA)}
Self-attention\cite{vaswani2017attention} has been widely used in various vision tasks\cite{liu2021swin,esser2021taming} due to its exceptional ability to understand global context. However, in the field of real-time object detection, the high computational cost and memory usage associated with self-attention calculations have posed continuous challenges. In YOLOv10\cite{wang2024yolov10}, partial self-attention was proposed to address this issue. They divided the features of the entire channel into two equal parts after the 1×1 convolution and fed only one part into the attention computation block. We enhance this approach by proposing the frequency separable self-attention (FSSA) module using the CFPNet structure, as shown in \cref{fig:architecture}.(c). Initially, the input feature map is split into low-frequency and high-frequency parts through octave convolution. Subsequently, only the low-resolution low-frequency feature maps are fed into the self-attention computation block consisting of multi-head self-attention (MHSA) and feed-forward network (FFN). These two parts are then connected and fused again through octave convolution. Additionally, we replace LayerNorm\cite{ba2016layer} in MHSA with BatchNorm\cite{ioffe2015batch} to speed up inference and to prevent excessive overhead due to the quadratic computational complexity, placing it only behind the last stage of the lowest-resolution backbone. By performing operations solely on the low-resolution low-frequency feature maps, our FSSA module integrates global representation capabilities into the model at a very low cost, significantly enhancing performance. Due to the nature of low-frequency feature maps, which primarily capture global information, the low-resolution feature maps substantially reduce the complexity of the MHSA operations, which have quadratic computational requirements, without significantly impacting the performance of the MHSA operations in extracting global information.

\subsection{Depthwise separable downsampling}
Typically, YOLO uses standard 3x3 convolutions with a stride of 2 in the backbone and neck to simultaneously decrease resolution and increase channel count. Similar to the depthwise bottleneck used in the FSB, we have implemented a more efficient downsampling layer by separating the tasks of resolution reduction and channel increase through depthwise separable convolutions. This method preserves information during downsampling while reducing latency.

\section{Experiment}

\subsection{Implementation Details}

We train the network for about 500 epochs using training and validation datasets from COCO\cite{lin2014microsoft}. The SGD optimizer is used. The initial learning rate is 1E−2, the final learning rate is 1E-5 with cosine scheduler, and the weight decay is 5E−3. A momentum is 0.937. The training runs 500 epochs with a batch size of 64, and the images inputted into the network were rescaled to 640×640.
In the experiment, we used Ubuntu 20.04.1 as the operating system with Python 3.10, PyTorch 2.2.0, and Cuda 11.7 as the desktop computational software environment. The experiment utilized NVIDIA RTX 3090 graphics cards as hardware. The implementation code of the neural network was modified based on the Ultralytics 8.1.24 version. The latencies of all models are tested on NVIDIA RTX 3090 GPU with ONNX runtime\cite{onnxruntime}.
\subsection{Comparison in resolution}
    
\begin{table}
\centering
\scriptsize
    \caption{Comparison of latency (ms) between Octave-YOLO-N and YOLOv8-N at different resolutions.}
    \begin{tabular}{c|cc|c}
      \toprule
      Resolution~ & ~YOLOv8-N~ & ~Octave-YOLO-N~ & ~Reduction\\
      \midrule
      320   & 1.71 & 1.68 & -1.75\% \\
      512   & 3.54 & 3.23 & -8.76\% \\
      640   & 5.38 & 4.34 & -19.33\% \\
      720   & 7.27 & 5.73 & -21.18\% \\
      1080  & 10.30& 6.59 & -36.02\% \\
    \bottomrule
    \end{tabular}
    \label{tab:resolution}
\end{table}
\cref{tab:resolution} presents a comparison of latencies between Octave-YOLO and YOLOv8 at various resolutions. When the input image is at a lower resolution of 320, the latency differences between Octave-YOLO and YOLOv8 are minimal. However, as the resolution of the input image increases, the disparity in latency grows. Particularly at a resolution of 1080, Octave-YOLO shows a significant difference, with over 3.7ms in reduced latency. This is attributed to Octave-YOLO's CFPNet-centered design, which enables it to handle complex computations at lower resolutions more efficiently.

\subsection{Comparison with state-of-the-arts}
\begin{table}
  \caption{Comparisons with state-of-the-arts.}
  \label{tab:coco}
  \scriptsize
  \centering
  \begin{tabular}{lcccccc}
    \toprule
    Model & \#Param.(M) & FLOPs(G) & AP$^{val}$(\%) \\
    \hline
    \specialrule{0em}{0.8pt}{0.8pt}
    YOLOv8-N \cite{Jocher_Ultralytics_YOLO_2023}     & 3.2   & 8.7      & 37.3    \\
    YOLOv6-3.0-N \cite{li2023yolov6v3} & 4.7\textcolor{red}{(+46.8\%)}   & 11.4\textcolor{red}{(+31.0\%)}& 37.0\textcolor{red}{(-0.8\%)}      \\
    \textbf{Octave-YOLO-N (Ours)} & 1.8\textcolor{green}{(-43.7\%)} & 5.3\textcolor{green}{(-39.1\%)} & 37.5\textcolor{green}{(+0.5\%)}     \\
    \hline
    \specialrule{0em}{0.8pt}{0.8pt}
    YOLOv8-S \cite{Jocher_Ultralytics_YOLO_2023}     & 11.2  & 28.6      & 44.9      \\
    YOLOv6-3.0-S \cite{li2023yolov6v3} & 18.5\textcolor{red}{(+65.1\%)}  & 45.3\textcolor{red}{(+58.4\%)}& 44.3\textcolor{red}{(-1.3\%)}   \\
    Gold-YOLO-S~\cite{wang2024gold}  & 21.5\textcolor{red}{(+91.9\%)}  & 46.0\textcolor{red}{(+60.8\%)}    & 45.4\textcolor{green}{(+1.1\%)}    \\
    RT-DETR-R18~\cite{zhao2023detrs} & 20.0\textcolor{red}{(+78.6\%)} & 60.0\textcolor{red}{(+109.8\%)} & 46.5\textcolor{green}{(+3.6\%)} \\
    \textbf{Octave-YOLO-S (Ours)} & 6.7\textcolor{green}{(-40.2\%)} & 16.5\textcolor{green}{(-42.3\%)} & 45.0\textcolor{green}{(+0.2\%)}    \\
    \hline
    \specialrule{0em}{0.8pt}{0.8pt}
    YOLOv8-M \cite{Jocher_Ultralytics_YOLO_2023}     & 25.9  & 78.9      & 50.6      \\
    YOLOv6-3.0-M \cite{li2023yolov6v3} & 34.9\textcolor{red}{(+34.7\%)}  & 85.8\textcolor{red}{(+8.7\%)}& 49.1\textcolor{red}{(-2.9\%)}  \\
    Gold-YOLO-M~\cite{wang2024gold}  & 41.3\textcolor{red}{(+59.5\%)}  & 87.5\textcolor{red}{(+10.9\%)}    & 49.8\textcolor{red}{(-1.6\%)}   \\
    RT-DETR-R34~\cite{zhao2023detrs} & 31.0\textcolor{red}{(+19.7\%)} & 92.0\textcolor{red}{(+16.6\%)} & 48.9\textcolor{red}{(-3.4\%)}   \\
    RT-DETR-R50m~\cite{zhao2023detrs} & 36.0\textcolor{red}{(+38.9\%)} & 100.0\textcolor{red}{(+26.7\%)} & 51.3\textcolor{green}{(+1.4\%)}   \\
    \textbf{Octave-YOLO-M (Ours)} & 12.0\textcolor{green}{(-53.7\%)} & 43.0\textcolor{green}{(-45.5\%)} & 49.7\textcolor{red}{(-1.8\%)}     \\
    \hline
    \specialrule{0em}{0.8pt}{0.8pt}
    YOLOv8-L \cite{Jocher_Ultralytics_YOLO_2023}     & 43.7  & 165.2      & 52.9  \\
    YOLOv6-3.0-L \cite{li2023yolov6v3} & 59.6\textcolor{red}{(+36.4\%)}  & 150.7\textcolor{green}{(-8.8\%)}& 51.8\textcolor{red}{(-2.1\%)}\\
    Gold-YOLO-L~\cite{wang2024gold}  & 75.1\textcolor{red}{(+71.8\%)}  & 151.7\textcolor{green}{(-8.2\%)}    & 51.8\textcolor{red}{(-2.1\%)} \\
    RT-DETR-R50~\cite{zhao2023detrs} & 42.0\textcolor{green}{(-3.9\%)} & 136.0\textcolor{green}{(-17.7\%)} & 53.1\textcolor{green}{(+0.4\%)} \\
    \textbf{Octave-YOLO-L (Ours)} & 18.2\textcolor{green}{(-58.3\%)} & 89.9\textcolor{green}{(-45.6\%)} & 52.3\textcolor{red}{(-1.1\%)}    \\
    \hline
    \specialrule{0em}{0.8pt}{0.8pt}
    YOLOv8-X \cite{Jocher_Ultralytics_YOLO_2023} & 68.2 & 257.8 & 53.9   \\
    RT-DETR-R101~\cite{zhao2023detrs} & 76.0\textcolor{red}{(+11.4\%)} & 259.0\textcolor{red}{(+0.5\%)} & 54.3\textcolor{green}{(+0.7\%)} \\
    \textbf{Octave-YOLO-X (Ours)} & 28.4\textcolor{green}{(-58.4\%)} & 140.0\textcolor{green}{(-45.7\%)} & 52.9\textcolor{red}{(-1.8\%)}   \\
    \bottomrule
  \end{tabular}
  \vspace{-13pt}
\end{table}

As shown in \cref{tab:coco}, our Octave-YOLO achieved performance nearly identical to the state-of-the-art, but with shorter end-to-end latency and less computational cost across various model scales. We first compare Octave-YOLO with baseline models, YOLOv8. On N / S / M / L / X five variants, our Octave-YOLO achieves 43.75\% / 40.18\% / 53.67\% / 58.35\% / 58.36\% fewer parameters, 39.08\% / 42.31\% / 45.50\% / 45.55\% / 45.65\% less calculations with only -0.54\% / -0.22\% / 1.78\% / 1.13\% / 1.86\% accuracy drop. Furthermore, for the smaller-sized N and S models, there has been a slight improvement in accuracy. Compared with other YOLOs, Octave-YOLO also exhibits superior trade-offs between accuracy and computational cost.

With the advancement of photographic technology, it is not difficult for the general public to obtain images with resolutions of 4K or higher using drones or smartphones. However, due to memory and computational constraints for real-time object detection, most detectors downsample to smaller resolutions such as 640x640 or 320x320 for inference, which can damage the detailed parts of the image and negatively impact accuracy.
On the other hand, Octave-yolo preserves high-resolution feature maps while performing computations at lower resolutions, achieving acceptable inference speeds even at high resolutions.

\subsection{Model Analyses}

\subsubsection{Ablation study.}
We present the ablation results for Octave-YOLO-N in the \cref{tab3}. Initially, the introduction of the CFPNet with FSB and depthwise bottleneck leads to a reduction of 0.96ms in latency while maintaining a competitive performance of 36.5\% AP. Additionally, depthwise separable downsampling results in a reduction of 0.5M parameters and 1.0 GFLOPs, with a latency reduction of 0.25 ms, effectively demonstrating its impact. Furthermore, through the introduction of the FSSA module, we achieve a 0.2 AP improvement with only 0.1 GFLOPs and 0.17 ms of overhead, showcasing performance nearly identical to the original YOLOv8.
\begin{table}[]
\caption{Ablation study with Octave-YOLO-N on COCO.}
\label{tab:ablation}
\scriptsize
\centering
\setlength{\tabcolsep}{2.5pt}
\begin{tabular}{lccc|ccccc}
    \toprule
    \# & FSB & DWDown & FSSA & \#Param.(M) & FLOPs(G) & AP$^{val}$(\%) & Latency(ms)  \\ 
    \hline
    \specialrule{0em}{0.8pt}{0.8pt}
    1  & & & & 3.2  &  8.7   &  37.3 & 5.38  \\ 
    2  & \checkmark  & &  &   2.1  &  6.1 & 36.5 & 4.42 \\
    3  & \checkmark & \checkmark &   &  1.6 & 5.1  & 36.2 & 4.17 \\
    4  & \checkmark & \checkmark & \checkmark & 1.8 & 5.2 & 37.5 & 4.34 \\
    \midrule
    \label{tab3}
\end{tabular}
\end{table}

\subsubsection{Analyses for depthwise bottleneck.}
In \cref{tab:ds}, we compare the effects of the depthwise bottleneck introduced as a computational block in the FSB with the conventional bottleneck used in C2f. Using the depthwise bottleneck results in a loss of 0.3 AP compared to the conventional bottleneck, but achieves an improvement of 44 ms in latency. We utilize the depthwise bottleneck to enhance the efficiency of our model.
\begin{table}[]    
\centering
    \begin{minipage}[t]{.49\linewidth}
    \caption{DWBottleneck.}
    \label{tab:ds}
    \scriptsize
    \centering
    \setlength{\tabcolsep}{2pt}
    \begin{tabular}{llcccccc}
        \toprule
        Model         & AP$^{val}$ & FLOPs(G) & Latency(ms) \\ 
        \hline
        \specialrule{0em}{0.8pt}{0.8pt}
        w/o DW & 37.8 & 6.5 & 4.78\\ 
        ours & 37.5 & 5.2 & 4.34\\
        \bottomrule
    \end{tabular}
    \end{minipage}
    \begin{minipage}[t]{.49\linewidth}
    \caption{Results of FSSA}
    \label{tab:fssa}
    \scriptsize
    \centering
    \setlength{\tabcolsep}{3pt}
    \begin{tabular}{llcccccc}
        \toprule
        Model         & AP$^{val}$ & FLOPs(G) & Latency(ms)  \\
        \hline
        \specialrule{0em}{0.8pt}{0.8pt}
        base & 37.3 & 8.7 & 5.38 \\
        \texttt{Trans.} & 38.6 & 9.8 & 6.13\\
        ours & 38.5 & 8.9 & 5.62\\
        \bottomrule
    \end{tabular}
    \end{minipage}
    
\end{table}
\subsubsection{Analyses for FSSA.}
We introduce FSSA to enhance performance by incorporating global modeling capabilities at a minimal additional cost. We first verify its effectiveness based on Octave-YOLO-N. Specifically, we initially introduce a conventional transformer block, which uses all feature maps without frequency domain splitting via CFPNet, consisting of MHSA followed by FFN, and denote it as \texttt{Trans.} As shown in \cref{tab:fssa}, while the conventional transformer block shows slightly better performance than FSSA, it comes with an unmanageable overhead of 0.75 ms. In contrast, FSSA achieves an enhancement in model capability with a negligible latency increase of 0.24 ms, while maintaining high efficiency.

\subsection{Visualization}
\begin{figure}[]
  \centering
  \begin{subfigure}{\linewidth}
  \centering
  \includegraphics[height=2.8cm]{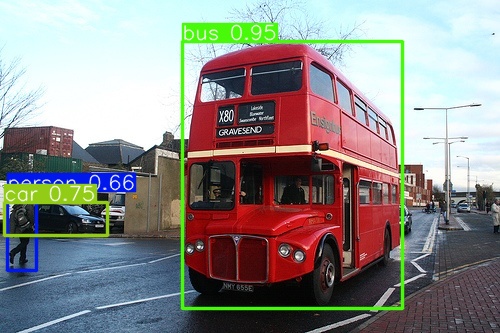}
  \includegraphics[height=2.8cm]{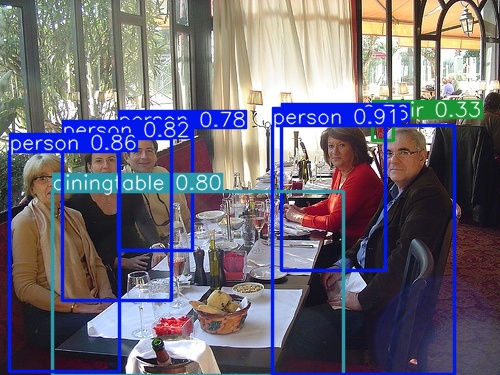}
  \includegraphics[height=2.8cm]{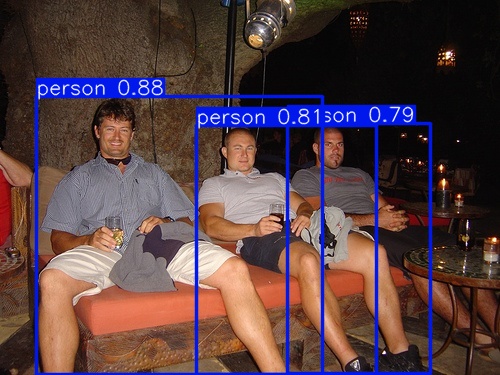}
  \caption{YOLOv8}
  \end{subfigure}
  \hfill
  \begin{subfigure}{\linewidth}
  \centering
  \includegraphics[height=2.8cm]{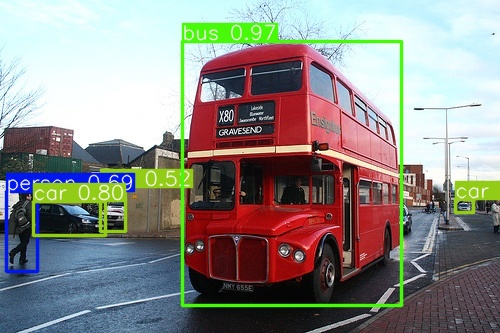}
  \includegraphics[height=2.8cm]{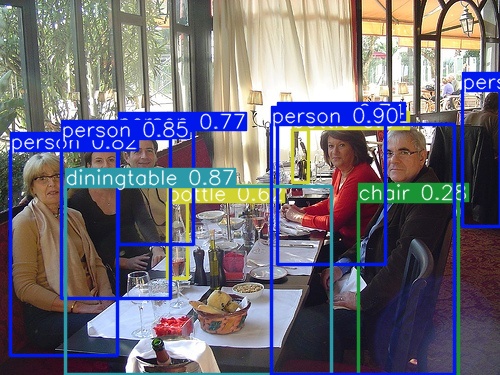}
   \includegraphics[height=2.8cm]{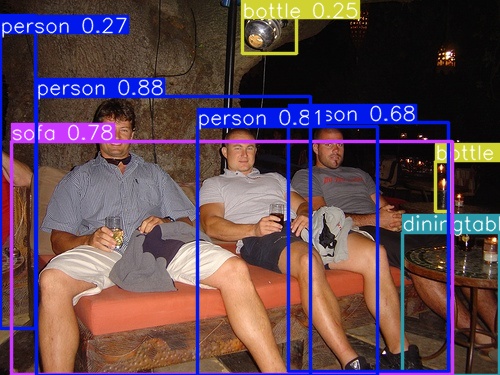}
  \caption{Octave-YOLO(ours)}
  \end{subfigure}
  \caption{Comparing image inference and visualization between YOLOv8-S and Octave-YOLO-S.}
  \label{fig:vis1}
\end{figure}

This analysis compares the object detection performance of YOLOv8-S and Octave-YOLO-S using images extracted from the COCO test set. We investigated variations across different scenes and object categories.

As illustrated in left side of \cref{fig:vis1}, a bus is parked on a driveway, with a car visible in the distance on the right. YOLOv8 failed to detect the distant car, whereas Octave-YOLO successfully identified it.

In middle, multiple individuals are seated around a table, with overlapping bounding boxes of both people and objects. Octave-YOLO excelled in detecting the overlapping bottles, while YOLOv8 demonstrated proficiency in identifying people but struggled with objects featuring overlapping boxes.

Moving to right, a nighttime photograph reveals YOLOv8's limitations in detecting objects beyond individuals in low-light conditions. Conversely, Octave-YOLO exhibited robust performance in identifying objects such as couches, dining tables, and bottles despite the challenging lighting environment.

\section{Conclusion}
In this paper, we introduce Octave-YOLO, a multi-scale object detector capable of processing high-resolution images in real time on embedded devices with limited memory and computational resources. We first propose the CFPNet to split the input feature map into high-frequency and low-frequency parts. We then process complex computations in the low-resolution, low-frequency areas and combine them with the preserved high-frequency areas, reducing redundant calculations and enhancing efficiency while alleviating the load of high-resolution input images. Additionally, we design building blocks and self-attention modules that apply the CFPNet, significantly increasing efficiency with minimal loss in accuracy. Extensive experiments demonstrate that Octave-YOLO achieves comparable performance and significantly reduced latency compared to other advanced detectors, effectively proving its superiority.


%
%
\bibliographystyle{splncs04}
\bibliography{main}
\end{document}